\pdfoutput=1

\documentclass[11pt]{article}

\usepackage{ACL2023}

\usepackage{times}
\usepackage{latexsym}

\usepackage[T1]{fontenc}

\usepackage[utf8]{inputenc}

\usepackage{microtype}

\usepackage{inconsolata}
\usepackage{graphicx}
\usepackage{subfigure}
\usepackage{booktabs}

\usepackage{multirow}
\usepackage{tikz}
\usepackage{pgfplots}
\pgfplotsset{compat=1.3}
\usepackage{pgfplotstable}
\usepackage{makecell}
\usepackage{multirow}
\usepackage{makecell}
\usepackage{braket}
\usepackage{amssymb}
\usepackage{amsmath}

\title{To Copy Rather Than Memorize: A Vertical Learning Paradigm for Knowledge Graph Completion}

\author{\textbf{Rui Li$^{1}$\footnotemark[1]\,\:, Xu Chen$^{2}$\footnotemark[2]\,\,, Chaozhuo Li$^{3}$\footnotemark[2]\,\,, Yanming Shen$^{1}$, Jianan Zhao$^{4}$,} \\
\textbf{Yujing Wang$^{5}$, Weihao Han$^{5}$, Hao Sun$^{5}$, Weiwei Deng$^{5}$, Qi Zhang$^{5}$, Xing Xie$^{3}$} \\
$^1$Dalian University of Technology,
$^2$Renmin University of China, \\
$^3$Microsoft Research Asia,
$^4$Université de Montréal, $^5$Microsoft \\
{\tt xu.chen@ruc.edu.cn},\quad{\tt cli@microsoft.com}
}

\begin{document}
\maketitle
\renewcommand{\thefootnote}{\fnsymbol{footnote}}
\footnotetext[1]{\,Work done during Rui Li's internship at MSRA.}
\footnotetext[2]{\,Correspondence to: Xu Chen <xu.chen@ruc.edu.cn>, Chaozhuo Li <cli@microsoft.com>.}
\renewcommand{\thefootnote}{\arabic{footnote}}
\begin{abstract}
Embedding models have shown great power in knowledge graph completion (KGC) task. \quad
By learning structural constraints for each training triple, these methods \emph{implicitly memorize} intrinsic relation rules to infer missing links.
However, this paper points out that the multi-hop relation rules are hard to be reliably memorized due to the inherent deficiencies of such implicit memorization strategy, making embedding models underperform in predicting links between distant entity pairs.
To alleviate this problem, we present 
Vertical Learning Paradigm (VLP), which extends embedding models by allowing to \emph{explicitly copy} target information from related factual triples for more accurate prediction.
Rather than solely relying on the implicit memory, VLP directly provides additional cues to improve the generalization ability of embedding models, especially making the distant link prediction significantly easier.
Moreover, we also propose a novel relative distance based negative sampling technique (ReD) for more effective optimization.
Experiments demonstrate the validity and generality of our proposals on two standard benchmarks.
Our code is available at \url{https://github.com/rui9812/VLP}.
\end{abstract}

\section{Introduction}
Knowledge graphs (KGs) structurally represent human knowledge as a collection of factual triples. Each triple $(h,r,t)$ represents that there is a relation $r$ between head entity $h$ and tail entity $t$. 
With the massive human knowledge, KGs facilitate a myriad of downstream applications \cite{xiong2017explicit}.
However, real-world KGs such as Freebase \cite{bollacker2008freebase} are far from complete~\cite{TransE}.
This motivates substantial research on the knowledge graph completion (KGC) task, i.e., automatically inferring missing triples.

As an effective solution for KGC, embedding model learns representations of entities and relations with pre-designed relation operations.
For example, TransE \cite{TransE} represents relations as translations between head and tail entities.
RESCAL~\cite{RESCAL}, DistMult \cite{yang2014embedding} and ComplEx~\cite{trouillon2016complex} model the three-way interactions in each triple.
RotatE \cite{sun2019rotate}, QuatE~\cite{QuatE} and DualE~\cite{DualE} represent relations as rotations in different dimensions.
Rot-Pro~\cite{Rot-Pro} further introduces the orthogonal projection for each relation.

\begin{figure}[t!]
\centering
\hspace{0.5mm}
\includegraphics[width=0.90\columnwidth]{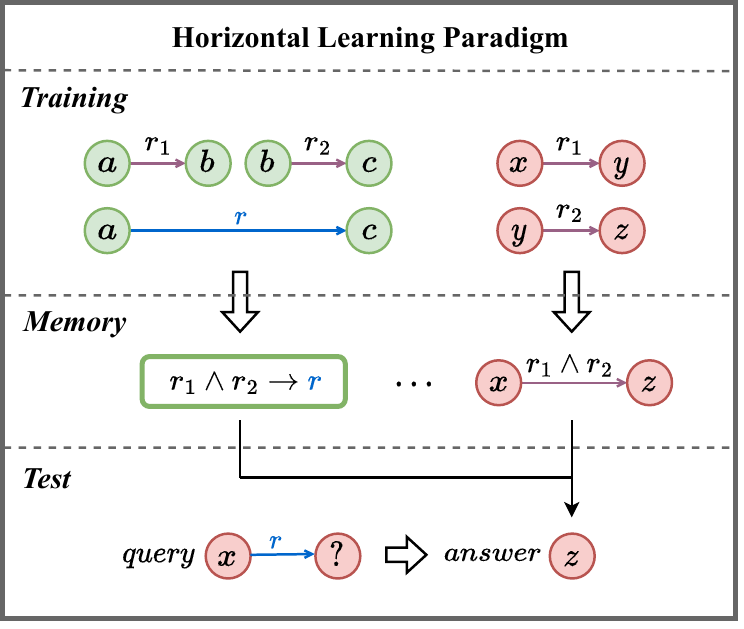}
\vspace{-1mm}
\caption{Learning paradigm of embedding models.}
\label{Fig-1}
\vspace{-4mm}
\end{figure}

\begin{table*}[t!]
\begin{center}
\begin{small}
\resizebox{0.85\textwidth}{!}{
\begin{tabular}{lcccccc}
\toprule
\multirow{3}{*}{Model} & \multirow{3}{*}{Score Function} & \multicolumn{4}{c}{$g(\boldsymbol{\rm {W}}_{r,1}\boldsymbol{\rm {h}} + \boldsymbol{\rm {b}}_r,\boldsymbol{\rm {W}}_{r,2}\boldsymbol{\rm {t}})$}   & \multirow{3}{*}{Space}                                           \\ \cmidrule(r){3-6}
\multicolumn{1}{c}{}                       & \multicolumn{1}{c}{}                                & $\boldsymbol{\rm {W}}_{r,1}$ & $\boldsymbol{\rm {b}}_r$ & $\boldsymbol{\rm {W}}_{r,2}$ & $g(\boldsymbol{\rm {q}},\boldsymbol{\rm {k}})$  & \multicolumn{1}{c}{}\\
\midrule
RESCAL \cite{RESCAL}                          &  $\boldsymbol{\rm {h}}^\top \boldsymbol{\rm {W}}_r \boldsymbol{\rm {t}}$           & $\boldsymbol{\rm {I}}$                                                   &  $\boldsymbol{\rm {0}}$         & $\boldsymbol{\rm {W}}_r$    & $\boldsymbol{\rm {q}}^\top \boldsymbol{\rm {k}}$          &    $\mathbb{R}$                                           \\ \addlinespace[0.5mm]
TransE \cite{TransE}                                     & $-\Vert \boldsymbol{\rm{h}}+\boldsymbol{\rm {r}}-\boldsymbol{\rm {t}} \Vert$    & $\boldsymbol{\rm {I}}$                                              & $\boldsymbol{\rm {r}}$          & $\boldsymbol{\rm {I}}$    & $-\Vert \boldsymbol{\rm {q}} - \boldsymbol{\rm {k}} \Vert$          & $\mathbb{R}$                                            \\ \addlinespace[0.5mm]
TransR \cite{lin2015learning}                                   & $-\Vert \boldsymbol{\rm {W}}_r\boldsymbol{\rm{h}}+\boldsymbol{\rm {r}}- \boldsymbol{\rm {W}}_r\boldsymbol{\rm {t}} \Vert$    & $\boldsymbol{\rm {W}}_r$                                              & $\boldsymbol{\rm {r}}$          & $\boldsymbol{\rm {W}}_r$    & $-\Vert \boldsymbol{\rm {q}} - \boldsymbol{\rm {k}} \Vert$          & $\mathbb{R}$                                            \\ \addlinespace[0.5mm]
DistMult \cite{yang2014embedding}                                  & $\boldsymbol{\rm {h}}^\top \mathrm{diag}(\boldsymbol{\rm {r}})\boldsymbol{\rm {t}}$              & $\mathrm{diag}(\boldsymbol{\rm {r}})$                                   &   $\boldsymbol{\rm {0}}$        & $\boldsymbol{\rm {I}}$    & $\boldsymbol{\rm {q}}^\top \boldsymbol{\rm {k}}$          &       $\mathbb{R}$                                        \\ \addlinespace[0.5mm]
ComplEx \cite{trouillon2016complex}                         & $\mathrm{Re}(\boldsymbol{\rm {h}}^\top \mathrm{diag}(\boldsymbol{\rm {r}})\boldsymbol{\rm {\overline{t}}})$                                                             & $\mathrm{diag}(\boldsymbol{\rm {r}})$          & $\boldsymbol{\rm {0}}$        & $\boldsymbol{\rm {I}}$     &  $\mathrm{Re}(\boldsymbol{\rm {q}}^\top \overline{\boldsymbol{\rm {k}}})$    & $\mathbb{C}$                                            \\ \addlinespace[0.5mm]
RotatE \cite{sun2019rotate}                      &     $-\Vert \boldsymbol{\rm {h}}\circ \boldsymbol{\rm {r}}-\boldsymbol{\rm {t}} \Vert$           &   $\mathrm{diag}(\boldsymbol{\rm {r}})$                                                 & $\boldsymbol{\rm {0}}$          & $\boldsymbol{\rm {I}}$    & $-\Vert \boldsymbol{\rm {q}} - \boldsymbol{\rm {k}} \Vert$          & $\mathbb{C}$    \\
\bottomrule
\end{tabular}}
\end{small}
\end{center}
\vspace{-3mm}
\caption{\label{Tab-1}
The score functions and GSF settings of several models, where $\circ$ denotes the Hadamard product.
}
\vspace{-4mm}
\end{table*}


Essentially, embedding models learn structural constraints for every factual triple during the training period.
For example, for each training triple $(h,r,t)$, TransE constrains that the head embedding $\boldsymbol{\rm {h}}$ plus the relation embedding $\boldsymbol{\rm {r}}$ equals the tail embedding $\boldsymbol{\rm {t}}$.
Such single-triple constraints empower embedding models to implicitly perceive (i.e., memorize) the high-order entity connections and intrinsic relation rules~\cite{sun2019rotate}.
As shown in Figure~\ref{Fig-1}, by imposing the structural constraints (e.g., $\mathbf{h}+\mathbf{r}=\mathbf{t}$ in TransE) on the five training triples,
embedding models can memorize the entity connection $(x,r_1\land r_2, z)$ and the relation rule $r_1\land r_2\rightarrow r$.
In this way, the missing link $(x,r,z)$ can be inferred at test time without any explicit prompt.
We refer to this single-triple learning paradigm as Horizontal Learning Paradigm (HLP), since the relation rules are implicitly induced by the horizontal paths between head and tail entities.
However, this paper shows that the HLP-based embedding models are hard to reliably memorize the multi-hop relation rules, which is attributed to inevitable single-triple bias and high-demanding memory capacity.
The unreliable multi-hop relation rules in the implicit memory cannot serve as rational basis for prediction, leading to the inferior performance of embedding models in predicting links between distant entity pairs.
This brings us a question: \emph{is there a general paradigm for embedding models to alleviate this problem of HLP and achieve superior performance?}

We give an affirmative answer by presenting Vertical Learning Paradigm (VLP), which endows embedding models with the ability to explicitly consult related factual triples (i.e., vertical references) for more accurate prediction.
Specifically, to answer $(h,r,?)$, VLP first selects $N$ relevant reference queries in the training graph, and then treats their ground-truth entities as the reference answers for embedding models to jointly predict the target $t$.
This learning process can be viewed as an \emph{explicit copy} strategy, which is different from the \emph{implicit memorization} strategy of HLP, making it significantly easier to predict distant links.
Moreover, to effectively optimize the models, we further propose a novel Relative Distance based negative sampling technique (ReD), which can generate more informative negative samples and reduce the toxicity of false negative samples.
Note that VLP and ReD are both general techniques and can be widely applied to various embedding models.
Our contributions are summarized as follows:
\begin{itemize}
    \item We show that existing embedding models underperform in predicting links between distant entity pairs, since they are hard to reliably memorize the multi-hop relation rules.
    \item We present a novel learning paradigm named VLP, which can empower embedding models to leverage explicit references as cues for more accurate prediction.
    \item We further propose a new relative distance based negative sampling technique named ReD for more effective optimization.
    \item We conduct in-depth experiments on two standard benchmarks, demonstrating the validity and generality of the proposed techniques. 
\end{itemize}

\section{Preliminaries}
To elicit our proposal from a general paradigm perspective, we give a bird's eye view of existing embedding models in this section.
We first review the problem setup of KGC task.
Afterwards, we summarize a generalized score function of embedding models and describe how the models learn to predict new links (i.e., horizontal learning paradigm). 

\subsection{Problem Setup}
Given the entity set $\mathcal{E}$ and relation set $\mathcal{R}$, a knowledge graph can be formally defined as a collection of factual triples $\mathcal{D}=\{(h,r,t)\}$, in which head/tail entities $h,t\in \mathcal{E}$ and relation $r \in \mathcal{R}$. 
KGC task aims to infer new links by answering a query $(h,r,?)$ or $(?,r,t)$.
As an effective tool for this task, embedding model learns representations of entities and relations to measure each candidate's plausibility with a pre-designed score function.

\subsection{Generalized Score Function}
\label{GSF}
Based on a series of previous works \cite{RESCAL, TransE, TransH, lin2015learning, yang2014embedding, trouillon2016complex, sun2019rotate, Rotate3D, Rot-Pro}, we summarize a generalized score function (GSF) of embedding models.
To facilitate presentation, we only describe the query case of $(h, r, ?)$, while $(?, r, t)$ can be similarly conduced. 

Given a query $(h,r,?)$ and a candidate answer $t$, GSF first maps the head embedding $\boldsymbol{\rm {h}}\in \mathbb{X}^{d_e}$ 
to the query embedding $\boldsymbol{\rm {q}}\in \mathbb{X}^{d_r}$ with a relation-specific linear transformation:
\begin{equation}
\boldsymbol{\rm {q}}= \boldsymbol{\rm {W}}_{r,1}\boldsymbol{\rm {h}} + \boldsymbol{\rm {b}}_r, \label{qf}
\end{equation}
where $\mathbb{X}\in \{\mathbb{R}, \mathbb{C}\}$ is the embedding space, $d_e$ and $d_r$ are the embedding dimensions of entities and relations, $\boldsymbol{\rm {W}}_{r,1}\in \mathbb{X}^{d_r\times d_e}$ and $\boldsymbol{\rm {b}}_r\in \mathbb{X}^{d_r}$ denote the relation-specific projection matrix and bias vector.

Then, GSF uses another linear function to generate the answer embedding $\boldsymbol{\rm {k}}\in \mathbb{X}^{d_e}$ from the tail embedding $\boldsymbol{\rm {t}}\in \mathbb{X}^{d_e}$:
\begin{equation}
\boldsymbol{\rm {k}} = \boldsymbol{\rm {W}}_{r,2}\boldsymbol{\rm {t}}, \label{af}
\end{equation}
where $\boldsymbol{\rm {W}}_{r,2}\in \mathbb{X}^{d_r\times d_e}$ denotes the relation transformation matrix for tail projections.

Finally, the plausibility score of the triple $(h,r,t)$ is calculated by a similarity function $g$:
\begin{equation}
score = g(\boldsymbol{\rm {q}},\boldsymbol{\rm {k}}). \label{mf}
\end{equation}

By combining the above three steps, we formally define the generalized score function $f_{g}$ as follows:
\begin{equation}
f_{g}(h,r,t) = g(\boldsymbol{\rm {W}}_{r,1}\boldsymbol{\rm {h}} + \boldsymbol{\rm {b}}_r,\boldsymbol{\rm {W}}_{r,2}\boldsymbol{\rm {t}}). 
\label{gf}
\end{equation}
With different choices of $\boldsymbol{\rm {W}}_{r,1}$, $\boldsymbol{\rm {b}}_{r}$, $\boldsymbol{\rm {W}}_{r,2}$ and $g$, GSF can be instantiated as specific score functions of existing models.
Table \ref{Tab-1} exhibits several popular methods and their corresponding GSF settings.

\subsection{Horizontal Learning Paradigm}
\label{HLP}
With the pre-defined score functions, embedding models commonly follow the horizontal learning paradigm, which constructs the single-edge constraints to implicitly memorize high-order entity connections and intrinsic relation rules.

Take RotatE to process the triples in Figure \ref{Fig-1} as an example. 
By imposing the rotation constraints on three triples $(a,r_1,b)$, $(b,r_2,c)$ and $(a, r, c)$, RotatE is able to perceive a two-hop entity connection and further induce a two-hop relation rule:
\begin{equation}
\begin{split}
\label{ad}
\left\{
			\begin{aligned}
			&\boldsymbol{\rm {b}}=\boldsymbol{\rm {a}}\circ \boldsymbol{\rm {r}}_1  \\
			&\boldsymbol{\rm {c}}=\boldsymbol{\rm {b}}\circ \boldsymbol{\rm {r}}_2  \\
                &\boldsymbol{\rm {c}}=\boldsymbol{\rm {a}}\circ \boldsymbol{\rm {r}}  \\
			\end{aligned}
			\right.
   \Rightarrow
			\boldsymbol{\rm {r}}=\boldsymbol{\rm {r}}_1\circ \boldsymbol{\rm {r}}_2.
\end{split}
\end{equation}
Similarly, the high-order connection can also be captured by constraining $(x,r_1,y)$ and $(y,r_2,z)$:
\begin{equation}
\begin{split}
\label{wz}
\left\{
			\begin{aligned}
			&\boldsymbol{\rm {y}}=\boldsymbol{\rm {x}}\circ \boldsymbol{\rm {r}}_1  \\
                &\boldsymbol{\rm {z}}=\boldsymbol{\rm {y}}\circ \boldsymbol{\rm {r}}_2  \\
			\end{aligned}
			\right.
			\Rightarrow
			\boldsymbol{\rm {z}}=\boldsymbol{\rm {x}}\circ \boldsymbol{\rm {r}}_1\circ \boldsymbol{\rm {r}}_2.
\end{split}
\end{equation}
Finally, by combining Equation (\ref{ad}) and  (\ref{wz}), RotatE is capable of inferring the missing link $(x,r,z)$.

\section{Motivation}
\label{motivation}

The motive of our work originates from an observation that embedding models underperform in predicting links between distant entity pairs (refer to Appendix~\ref{obsresult} for more details).
Since the effectiveness of embedding models is largely determined by the ability to learn intrinsic relation rules \cite{sun2019rotate, Rot-Pro, HousE}, such inferior performance reveals that the models are hard to memorize the multi-hop relation rules.
We attribute this deficiency to the \emph{multi-hop bias accumulation} and \emph{high-demanding memory capacity} in the implicit memorization strategy of HLP.

\vspace{-1mm}
\paragraph{Multi-hop Bias Accumulation} 
The HLP-based embedding models implicitly perceive the multi-hop relation rules by constraining each training edge as shown in Section \ref{HLP}. 
Nevertheless, the single-edge constraints inevitably have biases during the optimization, which will accumulate with the increase of relation hops.
This bias accumulation makes the memorized relation rules unreliable, leading to the deficient generalization ability for link prediction between distant entities.
Concretely, considering the single-edge biases, the rule learning process in Equation (\ref{ad}) can be rewritten as:
\begin{align}
\vspace{-2mm}
\left\{
			\begin{aligned}
			&\boldsymbol{\rm {b}}=\boldsymbol{\rm {a}}\circ \boldsymbol{\rm {r}}_1\circ \boldsymbol{\rm {\epsilon}}_{1}  \\
			&\boldsymbol{\rm {c}}=\boldsymbol{\rm {b}}\circ \boldsymbol{\rm {r}}_2\circ \boldsymbol{\rm {\epsilon}}_{2}  \\
                &\boldsymbol{\rm {c}}=\boldsymbol{\rm {a}}\circ \boldsymbol{\rm {r}}\circ \boldsymbol{\rm {\epsilon}}_{0}\\
			\end{aligned}
			\right.
\Rightarrow
			\boldsymbol{\rm {r}}=\boldsymbol{\rm {r}}_1\circ \boldsymbol{\rm {r}}_2\circ 
            \boldsymbol{\rm {\epsilon}}_{abc},\label{mba}
\vspace{-2mm}
\end{align}
where $\boldsymbol{\rm {\epsilon}}_{abc}=\boldsymbol{\rm {\epsilon}}_{0}^{-1}\circ\boldsymbol{\rm {\epsilon}}_{1}\circ\boldsymbol{\rm {\epsilon}}_{2}$ is the cumulative bias.
Note that $\boldsymbol{\rm {\epsilon}}_{abc}$ is triple-dependent, which makes it intractable for other queries, e.g., $(x,r,?)$ in Figure \ref{Fig-1}, to rely on this rule for prediction. 
\vspace{-1mm}
\paragraph{High-demanding Memory Capacity} The HLP-based models essentially learn the general rules from the relation paths between head and tail entities.
With the increase of path length, the quantity of different paths (or rules) expands exponentially \cite{Pathcon}.
This requires intensive memory to memorize the whole crucial relation rules.
However, the modeling capacity of embedding models is insufficient to meet this requirement.
Since these models constrain basic edges to form long-range paths following the bottom-up design of HLP, they are more inclined to memorize the low-order rules and forget the high-order rules.

\vspace{-1mm}
\paragraph{Design Goal}
We seek to develop a general technique to alleviate the "\emph{Hard to Memorize}" problem of existing embedding models.

A straight-forward strategy is to directly extract and process the enclosing subgraph between head and tail entities \cite{GraIL}, which can avoid the multi-hop bias accumulation.
However, such a sophisticated procedure needs to be executed once for each candidate triple, which brings enormous training and test time costs. For example, GraIL \cite{GraIL} takes about 1 month to infer on the full FB15k-237 test set \cite{NBFNet}.
Moreover, the enclosing subgraph extraction is also constrained by the path length, severely harming the performance of link prediction.

Therefore, this paper aims to propose a general framework which can: (1) alleviate the deficiency of HLP; (2) enjoy the merits of validity and generality with tractable computational costs.

\begin{figure}[t!]
\centering
\hspace{0.5mm}
\includegraphics[width=0.90\columnwidth]{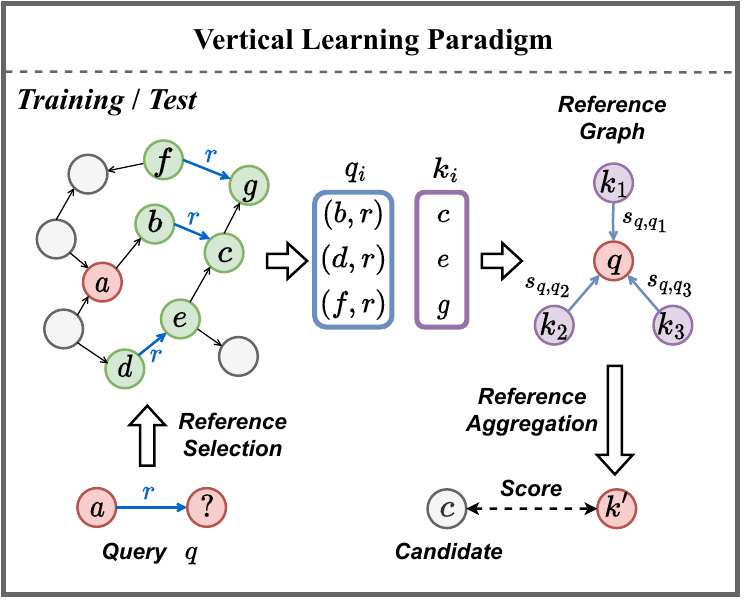}
\vspace{-2mm}
\caption{Vertical learning paradigm consisting of reference query selection, reference graph construction and reference answer aggregation.}
\label{VLP}
\vspace{-4mm}
\end{figure}

\section{Methodology}

\subsection{Vertical Learning Paradigm}
\label{method-VLP}
Inspired by the notion that “\emph{to copy is easier than to memorize}”~\cite{KNNLM},
we propose a vertical learning paradigm for KGC task.
Different from the implicit memorization strategy of HLP, VLP provides embedding models with the ability to reference related triples as cues for prediction, which can be viewed as an explicit copy strategy.

More concretely, we present the overall pipeline of VLP in Figure \ref{VLP}. Given a query $(h,r, ?)$, the procedure of predicting tail $t$ can be divided into reference query selection, reference graph construction and reference answer aggregation.

\vspace{-1mm}
\paragraph{Reference Query Selection}
For the input query $q=(h,r,?)$, the VLP-based models first select $N$ entity-relation pairs $(h_i, r)$ in the training graph as the reference queries $\{q_i\}_{i=1}^N$, which can provide relevant semantics for prediction.
For example, to answer (\emph{Jill Biden}, \emph{lives\_in}, ?), we can reference the answer-known query (\emph{Joe Biden}, \emph{lives\_in}, ?) for target information, since \emph{Joe Biden} and \emph{Jill Biden} are highly related.
One intuitive way for the reference selection is to choose the top-$k$ entities in terms of the cosine similarity between $h$ and all entities involved in relation $r$ during the optimization.
Nevertheless, this approach incurs high computational costs and is intractable. 
Numerically, the time complexity of such similarity calculation is $O(n_rd_e)$, where $n_r$ is the number of $r$-involved entities and $n_r\approx \left | \mathcal{E} \right | \gg d_e$ in the worst case.

In this work, inspired by the small world principle \cite{smallworld, GraphDist}, in which related individuals are connected by short chains (e.g., \emph{Joe Biden} and \emph{Jill Biden} are directly connected by the marriage relationship), we introduce the graph distance based approach for efficient reference query selection.
Specifically, we select $N$ $r$-involved entities $\{h\}_{i=1}^N$ closest to $h$ in terms of their relative graph distance (i.e., the shortest path length on the training graph).
The corresponding ground-truth targets $t_i$ of the reference queries $q_i = (h_i, r, ?)$ are referred as reference answers.
In this way, VLP-based models can pre-retrieve $N$ related references for every input query, thus incurring no additional computational cost for training and inference. 

\vspace{-1mm}
\paragraph{Reference Graph Construction}
After the efficient reference retrieval, we construct an edge-attributed reference graph to integrate the selected $N$ reference queries and their corresponding answers with the input query.
As shown in Figure \ref{VLP}, the input query $q$ is regarded as the central node, and the reference answers $t_i$ are treated as the $N$ neighbors.
VLP-based models aims to leverage the explicit reference answers for prediction.
However, since there is no guarantee that $t_i$ is the same as the target tail $t$, it is unreasonable to directly copy $t_i$ without any modification.
For example, to answer (\emph{England}, \emph{capital\_is}, ?), we cannot directly copy the answer of (\emph{France}, \emph{capital\_is}, ?). 

Therefore, we introduce the query similarity $s_{q,q_i}$ as the edge attribute between $q$ and $t_i$.
By considering the query differences, VLP-based models are able to adaptively copy the reference answers.
For example, to answer the input (\emph{England}, \emph{capital\_is}, ?), we can adjust the target information from \emph{Paris} in terms of the difference between (\emph{France}, \emph{capital\_is}, ?) and the input query.


\vspace{-1mm}
\paragraph{Reference Answer Aggregation}
With the constructed reference graph, VLP-based models learn to explicitly gather target information from neighbor answers for prediction.
Specifically, based on the generalized functions summarized in Section \ref{GSF}, the central node embedding $\mathbf{q}$ and neighbor node embedding $\mathbf{k}_i$ 
can be defined as:
\begin{equation}
\vspace{-2mm}
\begin{split}
\boldsymbol{\rm {q}}&= \boldsymbol{\rm {W}}_{r,1}\boldsymbol{\rm {h}} + \boldsymbol{\rm {b}}_r, \\
\boldsymbol{\rm {k}}_i&= \boldsymbol{\rm {W}}_{r,2}\boldsymbol{\rm {t}}_i.
\end{split}
\vspace{-2mm}
\end{equation}
The edge embedding $\mathbf{s}_{q,q_i}$ (i.e., query similarity embedding) can be further defined as:
\begin{equation}
\begin{split}
\boldsymbol{\rm {s}}_{q,q_i}&=\boldsymbol{\rm {q}}-\boldsymbol{\rm {q}}_i\\
&= \boldsymbol{\rm {W}}_{r,1}(\boldsymbol{\rm {h}}-\boldsymbol{\rm {h}}_i).
\end{split}
\end{equation}
Then, combining the neighbor nodes and edge attributes, VLP-based models aggregate the reference answers to generate the final embedding $\mathbf{t}'$:
\begin{equation}
\vspace{-2mm}
\begin{split}
\label{agg}
\mathbf{t}'&=\sigma(\mathbf{W}_{agg}[\mathbf{t}_N, \mathbf{q}]),\\
\mathbf{t}_N&= \frac{1}{N}\sum_{i=1}^{N}(\mathbf{W}_{node}\mathbf{k}_i+ \mathbf{W}_{edge}\mathbf{s}_{q,q_i}),   
\end{split}
\vspace{-2mm}
\end{equation}
where $\sigma(\cdot)$ is a nonlinear activation function (e.g., tanh), $[\cdot,\cdot]$ is the concatenate operation, $\mathbf{W}_{agg}$, $\mathbf{W}_{node}$ and $\mathbf{W}_{edge}$ are shared projection matrices.
The output $\mathbf{t}'$ should be close to the target tail embedding $\mathbf{t}$ in the latent space, whose score can be revealed by the cosine similarity:
\begin{equation}
    f_{c}(h,r,t)=\frac{\mathbf{t}^\top \mathbf{t}'}{\Vert \mathbf{t} \Vert \Vert \mathbf{t}' \Vert}
\end{equation}
We highlight that the VLP's aggregating strategy in Equation (\ref{agg}) differs from GNN-based methods \cite{COMPGCN, A2N, WGCN, RGCN}. For each query $(h,r,?)$, regardless of whether the reference query is a neighbor of $h$ in the training graph, VLP-based models  can directly attend to the reference answer throughout the entire training set.

\vspace{-1mm}
\paragraph{Score Function} For each triple $(h,r,t)$ in the test sets, to alleviate the deficiency of HLP and predict more accurately, we integrate the vertical score $f_c$ with the horizontal score $f_g$ to form the final score function $f$ with a weight hyper-parameter $\lambda$:
\begin{equation}
    f(h,r,t)=f_{c}(h,r,t)+\lambda f_g(h,r,t).
\end{equation}
Note that VLP can be widely applied to various embedding models, since the reference aggregation is designed on the generalized score function.

\vspace{-1mm}
\paragraph{Complexity Analysis}
Compared with the vanilla embedding models, the VLP-based models only bring a few additional parameters, i.e., the shared aggregation matrices in Equation (\ref{agg}).
Therefore, the VLP-based models have the same space complexity as the HLP-based models, i.e., $O(\left | \mathcal{E} \right |d_e)$.
In the aspect of time cost for processing single triple, the time complexity of vanilla embedding models is $O(d_rd_e)$, derived from the generalized score function in Equation (\ref{gf}).
The VLP-based models require the same computation for each reference, which produces the time complexity of $O(Nd_rd_e)$.
Such computation is tractable since a small $N$ (no more than $8$) is enough for VLP-based models to achieve high performance in the experiments.

\subsection{Optimization}
During training, we jointly optimize $f_c$ and $f_{g}$ by a two-component loss function with coefficient $\alpha$:
\begin{equation}
\vspace{-3mm}
    L=L_1+\alpha L_2.
\vspace{-3mm}
\end{equation}.

For the former one, we use the cross-entropy between predictions and labels as training loss:
\begin{equation}
\vspace{-1mm}
L_1=-\sum_{i=1}^{\left | \mathcal{E} \right |}y_i\log p_i,   
\label{l1}
\vspace{-1mm}
\end{equation}
where $p_i$ and $y_i$ are the $i$-th components of $\mathbf{p}$ and $\mathbf{y}$, respectively;
$\mathbf{p\in \mathbb{R}^{\left | \mathcal{E} \right |}}$ is calculated by applying the softmax function to the "1-to-All" \cite{CAN} results of $f_c$;
$\mathbf{y\in \mathbb{R}^{\left | \mathcal{E} \right |}}$ is the one-hop vector that indicates the position of true label.

For the later one, negative sampling has been proved quite effective in extensive works \cite{Rot-Pro,sun2019rotate}.
Formally, for a positive triple $(h,r,t)$, we first sample a set of entities $\{t'_i\}_{i=1}^{l}$ (or $\{h'_i\}_{i=1}^{l}$) based on the \emph{pre-sampling weights} $p_0$ to construct negative triples $(h,r,t'_i)$ (or $(h'_i,r,t)$). 
With these samples, a negative sampling loss is designed to optimize embedding models:
\begin{align}
L_2 = &- \sum_{i=1}^l p_1(h'_i,r,t'_i)\log{\sigma(-f(h'_i,r,t'_i)-\gamma)}\notag\\
&- \log{\sigma(\gamma+f(h,r,t))},\label{loss}
\end{align}
where $\gamma$ is a pre-defined margin, $\sigma$ is the sigmoid function, $l$ denotes the number of negative samples,  $(h'_i,r,t'_i)$ is a negative sample against $(h,r,t)$. Importantly, $p_1(h'_i,r,t'_i)$ is the \emph{post-sampling weight}, which determines the proportion of $(h'_i,r,t'_i)$ in the current optimization.

\begin{figure}[t!]
\centering
\includegraphics[width=0.875\columnwidth]{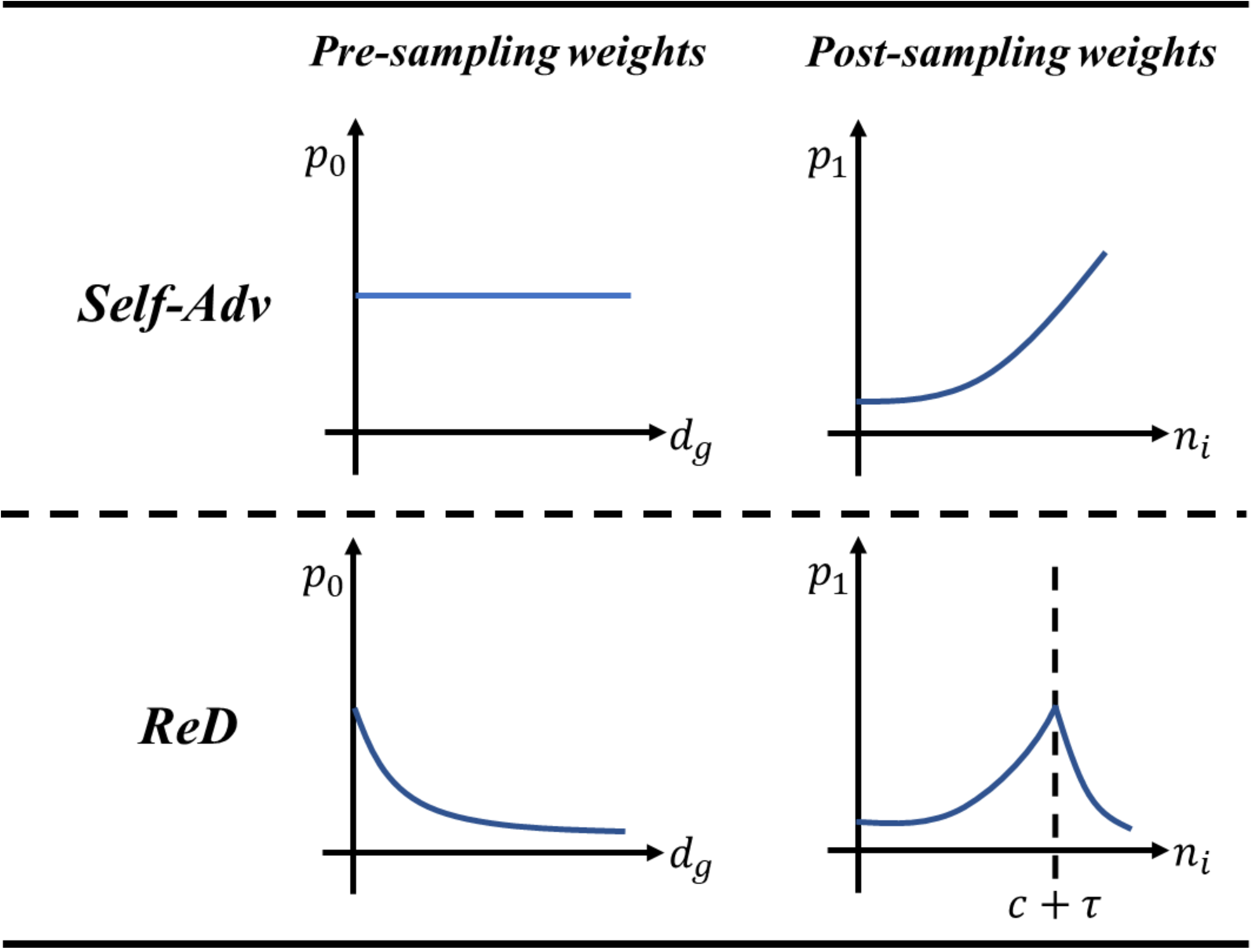}
\vspace{-1mm}
\caption{\label{redcomp}Comparison between ReD and Self-Adv.}
\label{ReD}
\vspace{-4mm}
\end{figure}

As shown in Figure~\ref{redcomp}, recent works \cite{Rot-Pro,PairRE,Rotate3D,sun2019rotate} utilize the self-adversarial technique (Self-Adv), in which the pre-sampling weights follow a uniform distribution and the post-sampling weights increase with the negative scores.
Differently, in this work, we propose a new approach named ReD based on the relative distance, which can draw more informative negative samples and reduce the toxicity of false negative samples.

For the pre-sampling weights, considering the deficiency of embedding models as described in Section \ref{motivation}, the distant entities are usually 
hard to be predicted as the target answer.
It reveals a rational priori, i.e., distant entities are more likely to form easy (meaningless) negative triples.
This inspires us to sample more hard (informative) negative triples based on the relative graph distance $d_g$.
As shown in Figure~\ref{redcomp}, the pre-sampling weight in ReD decreases with the increase of graph distance between head and tail entities. 
Formally, for a training query $(h, r, ?)$, we pre-sample entities $t'$ to construct negatives from the following distribution:
\begin{equation}
\label{preweight}
p_{0}(h,r,t')=\frac{\exp{-\alpha_0 d_g(h,t')}}{\sum_{i=1}^{\left | \mathcal{E} \right |}\exp{-\alpha_0 d_g(h,t'_i)}},
\end{equation}
where $\alpha_0$ is the pre-sampling temperature, $d_g(\cdot,\cdot)$ outputs the relative graph distance between two entities. 
Note that the calculation of $d_g(\cdot,\cdot)$ 
is a one-time preprocessing step, which will not bring additional training overhead.

\begin{table*}[t]
\begin{center}
\begin{small}
\resizebox{0.80\textwidth}{!}{
\begin{tabular}{lcccccccc}
\toprule
 & \multicolumn{4}{c}{WN18RR}                                                                                                                                                             & \multicolumn{4}{c}{FB15k-237}                                                                                                                                       \\ \cmidrule(r){2-5} \cmidrule(r){6-9}
 Model                                                     & MRR                                & H@1                                & H@3                                & H@10                                                             & MRR                                & H@1                                & H@3                                & H@10                                           \\
\midrule
TransE~\cite{TransE}$\dagger$                                               & .226                                 & -                                     & -                                     & .501                                                                  & .294                                 & -                                     & -                                     & .465                                 \\
ConvE~\cite{ConvE}                                                 & .43                                  & .40                                  & .44                                  & .52                                                                 & .325                                 & .237                                 & .356                                 & .501                \\
A2N~\cite{A2N}                                                & .45                              & .42                              & .46                             & .51                                                           & .317                             & .232                              & .348                              & .486                \\
QuatE~\cite{QuatE}                                              & .481                                 & .436                                 & .500                                 & .564                                                          & .311                                 & .221                                 & .342                                 & .495                         \\
CompGCN~\cite{COMPGCN}                                               & .479                              & .443                              & .494                              & .546                                                            & .355                              & \underline{.264}                              & .390                              & .535                       \\
PairRE~\cite{PairRE}                                                 & .455                              & .413                              & .469                              & .539                                                             & .348                              & .254                              & .384                              & .539                         \\
DualE~\cite{DualE}                                                 & .482                                 & .440                                 & .500                                 & .561                                                            & .330                                 & .237                                 & .363                                 & .518                         \\
Rot-Pro~\cite{Rot-Pro}                                                 & .457                                 & .397                                 & .482                                 & .577                                                                & .344                                 & .246                                 & .383                                 & .540                                   \\
CAKE~\cite{CAKE}                                                & -                              & -                              & -                             & -                                                               & .321                                 & .226                                 & .355                                 & .515                                   \\
REP~\cite{REP}                                                 & .488                                 & .439                                 & .505                                 & \textbf{.588}                                                               & .354                                 & .262                                 & .388                                 & .540                                   \\
ReflectE~\cite{ReflectE}                                                 & .488                                 & .450                                 & .501                                 & .559                                                                & \underline{.358}                                 & .263                                 & .396                                 & \textbf{.546 }                                  \\
\midrule
DistMult~\cite{yang2014embedding}$\diamond$                                             & .439                              & .392                                 & .453                                 & .534                                                          & .308                              & .220                                 & .337                                 & .485                            \\
DistMult-VLP  & .462 & .421 & .474 & .545  & .347 & .256 & .379 & .528 \\
\midrule
ComplEx~\cite{trouillon2016complex}$\diamond$                                               & .466                              & .423                              & .484                              & .552                                                             & .328                              & .235                              & .354                              & .511                    \\
ComplEx-VLP                    & \underline{.494} & \underline{.450} & \underline{.508} & .580  & .354 & .258 &  \underline{.396} & .536\\
\midrule
RotatE~\cite{sun2019rotate}                                                                               & .476                                 & .428                                 & .492                                 & .571                                                                  & .338                                 & .241                                 & .375                                 & .533                      \\
RotatE-VLP                    & \textbf{.498} & \textbf{.455} & \textbf{.514} & \underline{.582}  & \textbf{.362} & \textbf{.271} &  \textbf{.397} & \underline{.542}\\
\bottomrule
\end{tabular}}
\end{small}
\end{center}
\vspace{-3mm}
\caption{\label{mainresult}
Link prediction results on WN18RR and FB15k-237. Best results are in \textbf{bold} and second best results are \underline{underlined}. $[\dagger]$: Results are taken from~\cite{ConvKB}. $[\diamond]$: we re-evaluate DistMult and ComplEx based on the open source codes from~\cite{sun2019rotate}, achieving better results than those reported in the original papers.
}
\vspace{-3mm}
\end{table*}

For the post-sampling weights, Self-Adv assigns greater weights to high scoring negative triples in Equation (\ref{loss}), which makes the optimization focus more on hard negatives.
However, this monotonically increasing strategy ignores the issue of false negatives, since triples with higher scores are more likely to be correct.
A more rational posteriori is that the easy negatives are underscored and the false negatives are overscored.
In this work, we use the relative latent distance between the positive and negative samples to determine whether the negative score is too low or too high.
Specifically, ReD defines the post-sampling weights as a distribution that first rises and then falls as the negative score increases. 
As shown in Figure~\ref{redcomp}, if the negative score is significantly greater than (or less than) the positive score, this negative sample is more likely to be false (or easy), and thus be assigned a small weight in the Equation \ref{loss}.
Formally, based on the positive score $c=f_g(h,r,t)$ and negative score $n_i=f_g(h'_i,r,t'_i)$,  the post-sampling weight in ReD is defined as:
\begin{align}
&p_1(h'_i,r,t'_i)=\frac{\exp{w(h'_i,r,t'_i)}}{\sum_j\exp{w(h'_j,r,t'_j)}},\notag\\
&w(h'_i,r,t'_i)=\begin{cases} 
        \alpha_1n_i, &n_i\leq c+\tau\\
        \alpha_1c-\alpha_2m_i, &n_i>c+\tau 
       \end{cases},\notag\\
&m_i=n_i-c-\tau,\label{postweight}
\end{align}
where $\alpha_1$ and $\alpha_2$ are the post-sampling temperatures. By combining the sampling weights in Equation (\ref{preweight}) and (\ref{postweight}), ReD is able to generate and process higher quality negatives for optimization.


\section{Experiment}

\subsection{Experimental Setup}
\paragraph{Datasets} We evaluate our proposal on two widely-used benchmarks: WN18RR \cite{ConvE} and FB15k-237 \cite{toutanova2015observed}.
More details can be found in Appendix \ref{datasetsappx}.

\vspace{-1mm}
\paragraph{Baselines} To verify the effectiveness and generality of our proposal, we combine the proposed techniques with three representative embedding models DistMult~\cite{yang2014embedding}, ComplEx~\cite{trouillon2016complex} and RotatE~\cite{sun2019rotate}.
For performance comparison, we select a series of embedding models as baselines in Table \ref{mainresult}.

\vspace{-1mm}
\paragraph{Implementation Details} 
We fine-tune the hyperparameters with the grid search on the validation sets.
Please see Appendix~\ref{impd} for more details.

\subsection{Main Results}
The experimental results are reported in Table \ref{mainresult}. Compared to DistMult, ComplEx and RotatE, all three VLP-based versions achieve consistent and significant improvements on both datasets.
For example, on WN18RR and FB15k-237 datasets, RotatE-VLP outperforms RotatE with $2.2\%$ and $2.4\%$ absolute improvements in MRR, respectively.
Such obvious gains reveal that the vertical contexts generally inject valuable information into the embedding models for more accurate prediction.

Moreover, one can further see that ComplEx-VLP and RotatE-VLP perform competitively with the SOTA baselines.
Specifically, RotatE-VLP surpasses all the baselines in terms of most metrics over both datasets;
ComplEx-VLP also achieves promising performance on FB15k-237 compared with the baselines.
The superior performance further confirms the effectiveness of our proposal.


\begin{table}[t]
\begin{center}
\begin{small}
\resizebox{0.98\columnwidth}{!}{
\begin{tabular}{lcccc}
\toprule
Distance $d_{ht}$ & \makecell[c]{$1$\\$(47.7\%)$}              & \makecell[c]{$2$\\$(12.7\%)$}              & \makecell[c]{$3$\\$(29.3\%)$}                                      & \makecell[c]{$4$\\$(10.3\%)$}         \\
\midrule
DistMult                   & 0.971          & 0.331          & 0.293                                  & 0.039                                  \\
DistMult-VLP               & 0.989 & 0.345 & 0.328                         & 0.053                         \\
Relative Imp.                & \textbf{+1.9\%}         & \textbf{+4.2\%}         & \textbf{+11.9\%} & \textbf{+35.9\%} \\
\midrule
ComplEx                    & 0.979          & 0.367          & 0.396                                  & 0.058                                  \\
ComplEx-VLP                & 0.985 & 0.400 & 0.449                         & 0.102                         \\
Relative Imp.                & \textbf{+0.6\%}         & \textbf{+9.0\%}         &  \textbf{+13.4\%} & \textbf{+75.9\%} \\
\midrule
RotatE                     & 0.986          & 0.375          & 0.378                                  & 0.091                                  \\
RotatE-VLP                 & 0.991 & 0.391 & 0.456                         & 0.111                         \\
Relative Imp.                & \textbf{+0.5\%}         & \textbf{+4.3\%}         & \textbf{+20.6\%} & \textbf{+22.0\%}\\
\bottomrule
\end{tabular}}
\end{small}
\end{center}
\vspace{-3mm}
\caption{\label{distanceresult}
MRR on each distance split of WN18RR.
}
\vspace{-5mm}
\end{table}

\subsection{Fine-grained Performance Analysis}
\paragraph{Performance on Distance Splits} Table \ref{distanceresult} reports the performance of three VLP-based models on the distance splits defined in Appendix~\ref{obsresult}.
One can observe that: (1) the VLP-based embedding models outperform the vanilla models across all the distance splits; (2) the VLP models achieve greater relative improvement on the split with larger $d_{ht}$.
For example, as $d_{ht}$ increases from $1$ to $4$, RotatE-VLP achieves $0.5\%$, $4.3\%$, $20.6\%$ and $22.0\%$ relative improvements over RotatE on the MRR metric, respectively.
This reveals that the explicit vertical contexts can significantly alleviate the limitations of memory strategy in the embedding models.

\begin{table}[t]
\begin{center}
\begin{small}
\resizebox{\columnwidth}{!}{
\begin{tabular}{l|ccc}
\toprule
Relation Name   & RotatE  & QuatE  & RotatE-VLP  \\
\midrule
hypernym         & 0.154 & 0.172  & \textbf{0.191} \\
instance\_hypernym   & 0.324 & 0.362     & \textbf{0.376} \\
member\_meronym      & 0.255 & 0.236     & \textbf{0.269} \\
synset\_domain\_topic\_of    & 0.334 & 0.395     & \textbf{0.411} \\
has\_part            & 0.205 & 0.210     & \textbf{0.220} \\
member\_of\_domain\_usage     & 0.277 & 0.372    & \textbf{0.375} \\
member\_of\_domain\_region    & 0.243 & 0.140     & \textbf{0.391} \\
derivationally\_related\_form  & 0.957 & 0.952 & \textbf{0.958}    \\
also\_see           & 0.627 & 0.607  & \textbf{0.635}    \\
verb\_group                  & 0.968 & 0.930     & \textbf{0.968} \\
similar\_to             & 1.000 & 1.000     & \textbf{1.000} \\
\bottomrule
\end{tabular}}
\end{small}
\end{center}
\vspace{-3mm}
\caption{\label{fineresult}
MRR on each relation of WN18RR.}
\vspace{-5mm}
\end{table}

\vspace{-1mm}
\paragraph{Performance on Each Relation} To verify the modeling capacity of our proposal from a fine-grained perspective, we explore the performance of VLP-based models on each relation of WN18RR following  \cite{QuatE}. As shown in Table \ref{fineresult}, 
compared to RotatE and QuatE, RotatE-VLP surpasses them on all the 11 relation types, confirming that the explicit reference aggregation brings superior modeling capacity.

\vspace{-1mm}
\paragraph{Performance on Mapping Properties} 
Table \ref{RMP} exhibits the performance of our proposal on different relation mapping properties~\cite{sun2019rotate} in FB15k-237. 
We observe that RotatE-VLP consistently outperforms RotatE across all RMP types.
Such advanced performance owes to the powerful modeling capability of the explicit copy strategy.

\begin{table}[t]
\small
\begin{center}
\resizebox{0.75\columnwidth}{!}{
\begin{tabular}{c|l|cc}
\toprule
Task                             & RMPs & \multicolumn{1}{l}{RotatE}              & RotatE-VLP \\
\midrule
\multirow{4}{*}{\makecell[c]{Predicting\\Head\\(MRR)}} & 1-to-1                        & 0.498                                 & \textbf{0.504} \\
                                 & 1-to-N                        & 0.475                                       &   \textbf{0.478} \\
                                 & N-to-1                       & 0.088                                      & \textbf{0.126} \\
                                 & N-to-N                       & 0.260                                        & \textbf{0.286} \\
\midrule
\multirow{4}{*}{\makecell[c]{Predicting\\Tail\\(MRR)}} & 1-to-1                         & 0.490                       & \textbf{0.499} \\
                                 & 1-to-N                        & 0.071                          & \textbf{0.093} \\
                                 & N-to-1                        & 0.747                       & \textbf{0.770} \\
                                 & N-to-N                        & 0.367                       & \textbf{0.388} \\
\bottomrule
\end{tabular}}
\end{center}
\vspace{-3mm}
\caption{\label{RMP}
MRR on mapping properties in FB15k-237.
}
\vspace{-4mm}
\end{table}

\subsection{Impact of Reference Quantity} 
\begin{figure}[tbh!]
\centering
\hspace{-9mm}
\begin{tikzpicture}[font=\large, scale=0.65]
                \begin{axis}[
                    legend cell align={left},
                    legend style={nodes={scale=1.0, transform shape}},
                    xlabel style={font=\Large},
                    xlabel={\#reference $N$},
                    xtick pos=left,
                    tick label style={font=\large},
                    ylabel style={font=\Large},
                    ylabel={MRR},
                    xtick={0, 2, 4, 6, 8, 10},
                    xticklabels={$0$, $2$, $4$, $6$, $8$, $10$},
                    ytick={0.440, 0.455, 0.470, 0.485, 0.500},
                    yticklabels={$0.440$,$0.455$,$0.470$, $0.485$, $0.500$},
                    legend pos=south east,
                    ymajorgrids=true,
                    grid style=dashed
                ]
                \addplot[
                    color=purple,
                    dotted,
                    mark options={solid},
                    mark=diamond*,
                    line width=1.5pt,
                    mark size=2pt
                    ]
                    coordinates {
                    (0, 0.476)
                    (2, 0.490)
                    (4, 0.495)
                    (6, 0.497)
                    (8, 0.498)
                    (10, 0.498)
                    };
                    \addlegendentry{RotatE-VLP}
                \addplot[
                    color=blue,
                    dotted,
                    mark options={solid},
                    mark=*,
                    line width=1.5pt,
                    mark size=2pt
                    ]
                    coordinates {
                    (0, 0.466)
                    (2, 0.483)
                    (4, 0.489)
                    (6, 0.493)
                    (8, 0.494)
                    (10, 0.494)
                    };
                    \addlegendentry{ComplEx-VLP}
                \addplot[
                    color=brown,
                    dotted,
                    mark options={solid},
                    mark=triangle*,
                    line width=1.5pt,
                    mark size=2pt
                    ]
                    coordinates {
                    (0, 0.439)
                    (2, 0.456)
                    (4, 0.460)
                    (6, 0.461)
                    (8, 0.462)
                    (10, 0.461)
                    };
                    \addlegendentry{DistMult-VLP}
                \end{axis}
                \end{tikzpicture}
    \vspace{-1mm}
    \caption{\label{number}Impact of reference quantity on WN18RR.}
    \vspace{-4mm}
\end{figure}
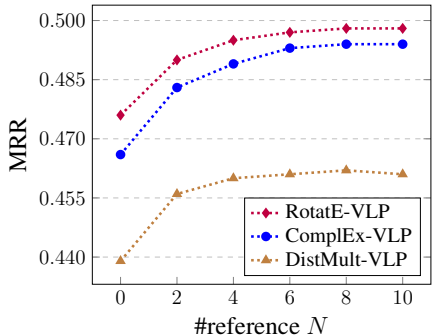

VLP aggregates target information from $N$ references pre-selected before training.
We investigate the impact of $N$ on the performance (MRR) of VLP-based models.
Figure \ref{number} shows the results on WN18RR dataset. 
As expected, all three VLP-based models with more vertical references achieve better performance than the ones with fewer references, since the aggregation of sufficient references brings the superior modeling capacity.
Moreover, we can observe that the models can achieve high performance with $N$ less than $10$, making the computation tractable as discussed in Section \ref{method-VLP}.

\subsection{Ablation Study of ReD}
To explore the effectiveness of the proposed ReD, we conduct ablation studies on the pre-sampling and post-sampling parts of the three VLP-based models. Table~\ref{ablation} shows the detailed results.
We can observe that the removal of any part reduces the performance, which demonstrates that ReD makes the model focus more on meaningful negative samples for more effective optimization.
Moreover, we also integrate ReD with original embedding models to verify the generality of this technique.
Please refer to Appendix~\ref{redablation} for more results. 

\section{Related Work}
Embedding models can be roughly categorized into distance based models and semantic matching models~\cite{PairRE}.

Distance based models 
use the Euclidean distance to measure the plausibility of each triple.
A series of work is conducted along this line, such as TransE~\cite{TransE} TransH~\cite{TransH}, TransR~\cite{lin2015learning}, RotatE~\cite{sun2019rotate}, PairRE~\cite{PairRE}, Rot-Pro~\cite{Rot-Pro}, ReflectE~\cite{ReflectE} and so on. 
TransE and RotatE are the most representative distance-based models, which represent relations as translations and rotations, respectively.
Semantic matching models utilize multiplicative functions to score each triple, including RESCAL~\cite{RESCAL}, DistMult~\cite{yang2014embedding}, ComplEx~\cite{trouillon2016complex}, QuatE~\cite{QuatE}, DualE~\cite{DualE} and so on.
Typically, RESCAL~\cite{RESCAL} 
defines each relation as the tensor decomposition matrix. 
DistMult~\cite{yang2014embedding} simplifies the relation matrices to be diagonal for preventing overfitting.
However, existing embedding models essentially follow the horizontal learning paradigm, underperforming in predicting links between distant entities.

Moreover, some advanced techniques are proposed to improve embedding models, such as graph encoders~\cite{RGCN,WGCN,COMPGCN,REP} and regularizers~\cite{N3}.
Note that our proposals are orthogonal to these techniques, and one can integrate them for better performance.

\begin{table}[t]
\small
\begin{center}
\resizebox{0.85\columnwidth}{!}{
\begin{tabular}{ccccc}
\toprule
                        & \multicolumn{2}{c}{WN18RR}                & \multicolumn{2}{c}{FB15k-237}             \\ \cmidrule(r){2-3} \cmidrule(r){4-5}
Model & MRR   &  H@10 & MRR   & H@10 \\
\midrule
DistMult-VLP            & 0.462 & 0.545                             & 0.347 & 0.528                             \\
\emph{w/o pre.}          & 0.456 & 0.537                             & 0.338 & 0.518                             \\
\emph{w/o post.}         & 0.458 & 0.542                             & 0.344 & 0.525                             \\
\midrule
ComplEx-VLP             & 0.494 & 0.580                             & 0.354 & 0.536                             \\
\emph{w/o pre.}          & 0.491 & 0.579                             & 0.344 & 0.529                             \\
\emph{w/o post.}         & 0.493 & 0.580                             & 0.345 & 0.531                             \\
\midrule
RotatE-VLP              & 0.498 & 0.582                             & 0.362 & 0.542                             \\
\emph{w/o pre.}          & 0.493 & 0.578                             & 0.355 & 0.540                             \\
\emph{w/o post.}         & 0.496 & 0.580                             & 0.359 & 0.539         \\
\bottomrule
\end{tabular}}
\end{center}
\vspace{-3mm}
\caption{\label{ablation}
Ablation study of ReD.
}
\vspace{-5mm}
\end{table}

\section{Conclusion}
In this paper, we present a novel learning paradigm named VLP for KGC task. 
VLP can be viewed as an explicit copy strategy, which allows embedding models to consult related triples for explicit references, making it much easier to predict distant links.
Moreover, we also propose ReD, a new negative sampling technique for more effective optimization. 
The in-depth experiments on two datasets demonstrate the validity and generality of our proposals.

\section*{Limitations}
Although our proposal enjoys the advantages of validity and generality, there are still two major limitations. 
First, VLP cannot directly generalize to the inductive setting, since VLP is defined based on the score functions of transductive embedding models.
One potential direction is to design an inductive reference selector for emerging entities.
Second, how to efficiently select more helpful references for prediction is still an open challenge.
We expect future studies to mitigate these issues.



\section*{Acknowledgements}
This work was supported in part by the National Natural Science Foundation of China under Grant 62276044, and also Sponsored by CAAI-Huawei MindSpore Open Fund.


\bibliography{anthology,custom}
\bibliographystyle{acl_natbib}

\appendix

\section{Experimental Observation}
\label{obsresult}
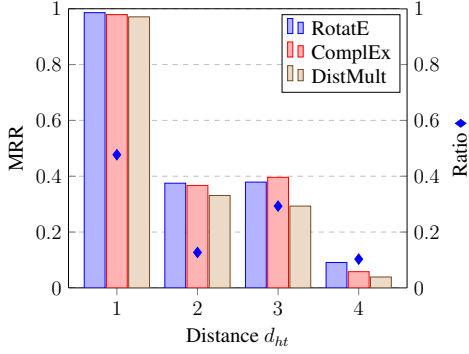
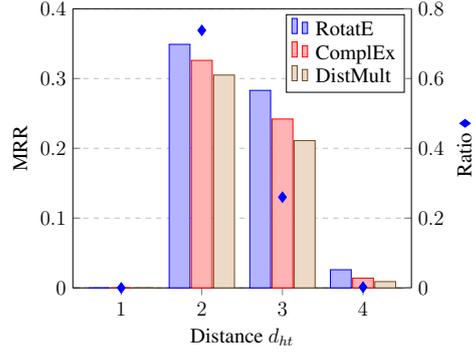
\begin{figure*}[t!]
\centering
\subfigure[MRR and ratio for each split in WN18RR]{\label{Fig-2a}
            \begin{tikzpicture}[font=\large,scale=0.65]
                \begin{axis}[
                    tick label style={font=\large},
                    ylabel style={font=\large},
                    axis y line*=left,
                    xtick pos=left,
                    ytick pos=left,
                    ylabel = MRR, 
                    enlarge x limits=0.20,
                    ymajorgrids=true,
                    xmajorgrids=false,
                    xtick={1, 2, 3, 4},
                    xticklabels={$1$, $2$, $3$, $4$},
                    xlabel={Distance $d_{ht}$},
                    grid style=dashed,
                    ybar=2*\pgflinewidth,
                    bar width = 12pt,
                    ymin=0,ymax=1.0,
                    ytick distance=0.2,
                    legend style={font=\large, column sep=2.5pt},
                    legend cell align={left},
                    xmin=1,xmax=4,
                    xtick distance=1,
                    ]
                \addplot coordinates {
                    (1, 0.986)
                    (2, 0.375)
                    (3, 0.379)
                    (4, 0.091)
                    };
                \addplot coordinates {
                    (1, 0.979)
                    (2, 0.367)
                    (3, 0.396)
                    (4, 0.058)
                    };
                \addplot coordinates {
                    (1, 0.971)
                    (2, 0.331)
                    (3, 0.293)
                    (4, 0.039)
                    };
                \legend{RotatE, ComplEx, DistMult}
                \end{axis}
                \begin{axis}[
                    tick label style={font=\large},
                    ylabel style={font=\large},
                    axis y line*=right,
                    axis x line=none,
                    ylabel={Ratio \ref{pgfplots:plot2}},
                    ymin=0,ymax=1.0,
                    ytick distance=0.2,
                    xmin=1,xmax=4,
                    xtick distance=1,
                    enlarge x limits=0.20,
                ]
                \addplot[color=blue,
                    only marks,
                    mark options={solid},
                    mark=diamond*,
                    mark size=3pt] coordinates {
                    (1, 0.477)
                    (2, 0.127)
                    (3, 0.293)
                    (4, 0.103)
                    };
                \label{pgfplots:plot2}
                \end{axis}
                \end{tikzpicture}
    }
\quad\quad\quad
\subfigure[MRR and ratio for each split in FB15k-237]{\label{Fig-2b}
            \begin{tikzpicture}[font=\large,scale=0.65]
                \begin{axis}[
                    tick label style={font=\large},
                    ylabel style={font=\large},
                    axis y line*=left,
                    xtick pos=left,
                    ytick pos=left,
                    ylabel = MRR, 
                    enlarge x limits=0.20,
                    ymajorgrids=true,
                    xmajorgrids=false,
                    xtick={1, 2, 3, 4},
                    xticklabels={$1$, $2$, $3$, $4$},
                    xlabel={Distance $d_{ht}$},
                    grid style=dashed,
                    ybar=2*\pgflinewidth,
                    bar width = 12pt,
                    ymin=0,ymax=0.4,
                    ytick distance=0.1,
                    legend style={font=\large, column sep=2.5pt},
                    legend cell align={left},
                    xmin=1,xmax=4,
                    xtick distance=1,
                    ]
                \addplot coordinates {
                    (1, 0.0)
                    (2, 0.349)
                    (3, 0.283)
                    (4, 0.026)
                    };
                \addplot coordinates {
                    (1, 0.0)
                    (2, 0.326)
                    (3, 0.242)
                    (4, 0.014)
                    };
                \addplot coordinates {
                    (1, 0.0)
                    (2, 0.305)
                    (3, 0.211)
                    (4, 0.009)
                    };
                \legend{RotatE, ComplEx, DistMult}
                \end{axis}
                \begin{axis}[
                    tick label style={font=\large},
                    ylabel style={font=\large},
                    axis y line*=right,
                    axis x line=none,
                    ylabel={Ratio \ref{pgfplots:plot1}},
                    ymin=0,ymax=0.8,
                    ytick distance=0.2,
                    xmin=1,xmax=4,
                    xtick distance=1,
                    enlarge x limits=0.20,
                ]
                \addplot[color=blue,
                    only marks,
                    mark options={solid},
                    mark=diamond*,
                    mark size=3pt] coordinates {
                    (1, 0.0)
                    (2, 0.738)
                    (3, 0.260)
                    (4, 0.002)
                    };
                \label{pgfplots:plot1}
                \end{axis}
                \end{tikzpicture}
    }
    \vspace{-3mm}
    \caption{The MRR results of three popular embedding models (DistMult, ComplEx and RotatE) tested on each distance split in WN18RR and FB15k-237 datasets. The blue diamond marks denote the ratio of each test split.}
    \label{Fig-2}
    \vspace{-2mm}
\end{figure*}
The motive of our work originates from an experimental observation, which shows that embedding models underperform in predicting links between distant entity pairs.
Specifically, according to the relative graph distance $d_{ht}$ between head and tail entities of each test triple, we divide the test sets of WN18RR and FB15k-237 into four splits. Three representative embedding models (DistMult, ComplEx and RotatE) are tested on each split.

Figure \ref{Fig-2} summarizes the detailed MRR results and split ratios on the two datasets.
We can observe that all three embedding models achieve promising results in link prediction between close entities, while the performance drops significantly in the prediction between distant entities.
For example, on the split where $d_{ht}=1$ in WN18RR, RotatE achieves excellent performance (MRR of 0.986), while on the split where $d_{ht}=2$, the performance of RotatE decreases by about 62\% (MRR of 0.375).



\section{Datasets}
\label{datasetsappx}
\begin{table}[ht]
\begin{center}
\begin{small}
\resizebox{0.65\columnwidth}{!}{\begin{tabular}{lcc}
\toprule
Dataset & WN18RR & FB15k-237 \\
\midrule
\#entity    & 40,943 & 14,541 \\
\#relation    & 11 & 237 \\
\#training  & 86,835 & 272,115 \\
\#validation    & 3,034 & 17,535 \\
\#test    & 3,134 & 20,466 \\
\bottomrule
\end{tabular}}
\end{small}
\end{center}
\vspace{-3mm}
\caption{\label{datasets}Statistics of five standard benchmarks.}
\end{table}
Table \ref{datasets} summarizes the detailed statistics of two benchmark datasets.
WN18RR~\cite{ConvE} and FB15k-237~\cite{toutanova2015observed} datasets are subsets of WN18~\cite{TransE} and FB15k~\cite{TransE} respectively with inverse relations removed. 
WN18 is extracted from WordNet~\cite{WordNet}, a database featuring lexical relations between words.
FB15k is extracted from Freebase~\cite{bollacker2008freebase}, a large-scale KG containing general knowledge facts.



\section{Implementation Details}
\label{impd}
\begin{table}[t]
\begin{center}
\begin{small}
\resizebox{0.75\columnwidth}{!}{\begin{tabular}{cc}
\toprule
Hyperparameter  & Search Space  \\
\midrule
$b$  & $\{256, 512, 1024\}$\\
$d$ &$\{500, 1000\}$\\
$\alpha_0, \alpha_1, \alpha_2$  & $\{0.1, 0.5, 1.0, 1.5\}$\\
$\lambda$  & $\{0.1, 0.3, 0.5, 0.7, 0.9\}$\\
$\gamma$  & $\{4, 6, 8, 11, 15\}$\\
\bottomrule
\end{tabular}}
\end{small}
\end{center}
\vspace{-3mm}
\caption{\label{Hyper_search}Hyperparameter search space.}
\vspace{-2mm}
\end{table}
We use Adam~\cite{Adam} as the optimizer and fine-tune the hyperparameters on the validation dataset. 
The hyperparameters are tuned by the grid search, including batch size $b$, embedding dimension $d$, negative sampling temperatures $\{\alpha_i\}_{i=0}^2$, loss weight $\lambda$ and fixed margin $\gamma$. The hyper-parameter search space is shown in Table \ref{Hyper_search}.

\section{Embedding Models with ReD}
\label{redablation}
\begin{table}[ht]
\small
\begin{center}
\resizebox{0.875\columnwidth}{!}{
\begin{tabular}{lcccc}
\toprule
                        & \multicolumn{2}{c}{WN18RR}                & \multicolumn{2}{c}{FB15k-237}             \\ \cmidrule(r){2-3} \cmidrule(r){4-5}
Model & MRR   & H@10 & MRR   & H@10 \\
\midrule
DistMult-Adv                & 0.439 & 0.534                             & 0.308 & 0.485                             \\
DistMult-ReD            & 0.445 & 0.539                             & 0.315 & 0.491                             \\
\midrule
ComplEx-Adv                 & 0.466 & 0.552                             & 0.328 & 0.511                              \\
ComplEx-ReD             &       0.470&                                   0.554&       0.335& 0.516                                  \\
\midrule
RotatE-Adv                  & 0.476 & 0.571                             & 0.338 & 0.533                             \\
RotatE-ReD              & 0.478 & 0.572                                  & 0.344      & 0.536          \\
\bottomrule
\end{tabular}}
\end{center}
\vspace{-3mm}
\caption{\label{kgered}
Results of DistMult, ComplEx and RotatE with different negative sampling techniques. X-Adv denotes the embedding model X combined with Self-Adv.
}
\vspace{-2mm}
\end{table}
To verify the generality of the proposed negative sampling technique ReD, we integrate ReD with three representative embedding models (i.e., DistMult, ComplEx and RotatE) for KGC task.
As shown in Table~\ref{kgered}, compared to Self-Adv, the embedding models combined with ReD achieve better performance on both datasets, since ReD guarantees more informative negative samples from both pre-sampling and post-sampling stages.

\end{document}